%% file: main.tex
\documentclass[10pt,twocolumn,letterpaper]{article}

\usepackage{wacv}              


\definecolor{wacvblue}{rgb}{0.21,0.49,0.74}
\usepackage[pagebackref,breaklinks,colorlinks,allcolors=wacvblue]{hyperref}

\def\wacvPaperID{819} 
\def\confName{WACV}
\def\confYear{2027}

\title{AGILE-GS: Anchor-Guided Fast Next-Best-View Selection for Active \\ 3D Gaussian Splatting\thanks{This work was supported in part by the ONR under grants numbers N00014-23-1-2779 and N00014-26-1-2246. \newline
$^\dagger$A.M. Khass and N. Motee are with the Department of Mechanical Engineering and Mechanics, Lehigh University, Bethlehem, PA, 18015, USA. {\tt\small \{ammb231,motee\}@lehigh.edu}.\endgraf}}

\author{Amirhossein Mollaei Khass$^\dagger$, Nader Motee$^\dagger$}

\begin{document}
\maketitle

\begin{abstract}
Radiance fields need hundreds of views, and their placement matters as much as their number. Next-best-view (NBV) selection for 3D Gaussian Splatting (3DGS) usually scores every candidate in the pool and keeps one. Searching for information and choosing a camera, however, are separable problems. We present AGILE-GS, an anchor-guided NBV method that separates the two. A virtual anchor pose is optimized on SE(3) by Riemannian gradient ascent on expected information gain. It need not be reachable or in the pool; it marks where the model is most uncertain. Candidates are scored against the anchor's viewing geometry, and a greedy ridge-leverage step distills the pool into a small, non-redundant shortlist without rendering any candidate. The shortlist can be used in two ways. AGILE-GS takes the first view on it as the next view, so no Fisher information is computed for any candidate. AGILE-GS+ computes the Fisher information gain of each shortlisted view and picks the best, so the expensive evaluation runs on a handful of views rather than the whole pool. On standard benchmarks and in closed-loop embodied acquisition, both match or exceed existing baselines while cutting selection latency by one to two orders of magnitude.
\end{abstract}




\section{Introduction}
\label{sec:introduction}
Volumetric radiance fields have become the dominant representation
for novel view synthesis and 3D reconstruction. 3D Gaussian
Splatting (3DGS)~\cite{kerbl20233d} represents a scene with an
explicit set of anisotropic Gaussian primitives. It matches the
fidelity of neural radiance fields and renders in real time.

\begin{figure}[t]
    \centering
    \includegraphics[
        width=0.9\textwidth,height=0.22\textheight,
        trim={11.5cm 5.22cm 2.5cm 3.6cm},
        clip
    ]{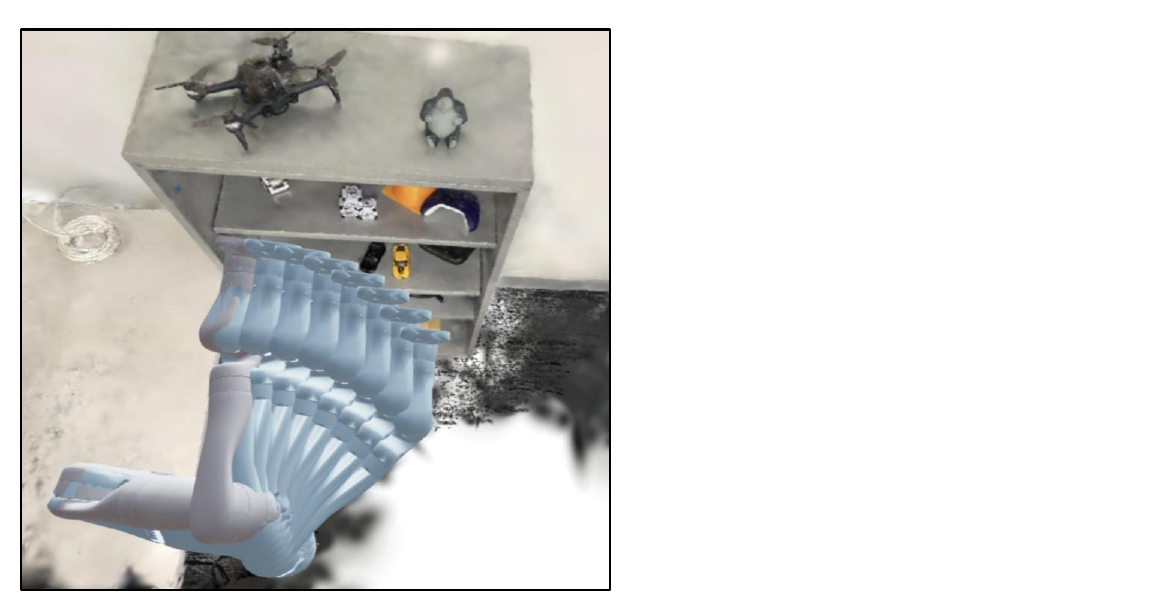}
    \vspace{-0.6 cm}
\caption{A simulated robot manipulator autonomously selects informative viewpoints for 3D Gaussian Splatting reconstruction.} 
\vspace{-0.6 cm}
\label{fig:autotrajectory}
\end{figure}

Deciding where a sensor should look next is a long-standing problem
in computer vision. It goes back to early work on active
perception~\cite{bajcsy1988active,aloimonos1988active} and
next-best-view planning~\cite{connolly1985determination}, where the
goal is to build a complete and accurate model from as few
observations as possible~\cite{scott2003view}. Radiance fields have
made this problem pressing again. 3DGS needs hundreds of posed views,
and reconstruction quality depends strongly on how those views are
distributed. Redundant or poorly placed observations leave parts of
the scene weakly constrained and produce geometric and photometric
artifacts~\cite{khass2025active,nagami2026vista,tao2024rtguiderealtimegaussiansplatting}.
For a robot, every additional capture also costs time and motion, so
views must be chosen rather than collected. Next-best-view (NBV)
selection addresses this by repeatedly choosing the observation
expected to most improve the current
model~\cite{chen2025activegamer,chen2026coverage,jiang2024fisherrf,
wilson2025pop}.

Existing NBV criteria trade fidelity against cost.
FisherRF~\cite{jiang2024fisherrf} and POp-GS~\cite{wilson2025pop}
estimate the expected uncertainty reduction directly in parameter
space. This is accurate, but it requires a differentiable
render--backward pass for every candidate, so the cost grows with
both the map size and the candidate pool.
COVER~\cite{chen2026coverage} avoids per-candidate rendering, but it
captures parameter uncertainty only indirectly. A further
inefficiency is shared by all of them. Densely sampled pools contain
many near-duplicate views, so a large fraction of any per-candidate
evaluation is spent telling apart observations that are effectively
redundant.

We take a different route. The most informative viewpoint may be
neither reachable nor present in the candidate pool, yet it can still
guide the choice among the views that are available. We therefore
replace per-candidate evaluation with a fast search for informative
candidates. A virtual anchor pose is first optimized on
$\mathrm{SE}(3)$ by Riemannian gradient ascent to maximize expected
information gain. This steers the anchor toward the region of pose
space that most reduces uncertainty in the current 3DGS model. Each
candidate then receives an anchor leverage score that measures how
well it reproduces the anchor's viewing geometry. A fast subset
selection step keeps the highest-scoring candidates and suppresses
those that are redundant with views already selected. The result is a
compact shortlist, obtained without rendering a single candidate.
\textsc{AGILE-GS} (\textbf{A}nchor-\textbf{G}uided Fast Next Best
V\textbf{i}ew Se\textbf{l}ection for Activ\textbf{e} 3D \textbf{G}aussian
\textbf{S}platting) returns the leading proposal from this shortlist directly.
No Fisher information is computed for any candidate, which gives the
lowest latency. \textsc{AGILE-GS+} instead reranks the shortlist by
expected information gain. It recovers model-aware accuracy at a
fraction of the full-pool cost, since the expensive evaluation runs
on a handful of views rather than all of them.

This work makes three contributions. First, we give a
continuous-to-discrete NBV formulation that optimizes a view pose on
$\mathrm{SE}(3)$ to maximize expected information gain and uses it to
guide discrete view selection. Second, we introduce an anchor
leverage score and a fast subset selection procedure that return a
compact, non-redundant proposal set without rendering any candidate.
Third, we present two algorithms, \textsc{AGILE-GS} and
\textsc{AGILE-GS+}, that pair anchor-guided geometric selection with
optional Fisher reranking over the shortlist. They match the
reconstruction accuracy of parameter-space methods at one to two
orders of magnitude lower selection latency. 
\vspace{-0.2 cm}
\section{Related Work}
\label{sec:related}

View planning has a long history in computer vision and robotics.
Connolly~\cite{connolly1985determination} used a partial octree model
to pick the view that uncovers the most unseen volume, and the survey
by Scott et al.~\cite{scott2003view} covers the model-based and
non-model-based methods that followed. These methods target range
sensors and explicit geometry. Radiance fields change the problem.
The model being improved is a set of continuous parameters rather
than an occupancy grid, so uncertainty has to be measured in that
parameter space.

Recent NBV methods do exactly this. ActiveNeRF~\cite{pan2022activenerf}
propagates predictive uncertainty through a NeRF.
FisherRF~\cite{jiang2024fisherrf} approximates the uncertainty
reduction with Fisher information. POp-GS~\cite{wilson2025pop} casts
3DGS acquisition as optimal experimental design. These objectives are
principled, but each candidate needs its own rendering or
differentiation pass, so the cost scales with both the model size and
the candidate pool. COVER~\cite{chen2026coverage} takes a geometric
route instead. It favors cameras that observe under-constrained
regions and scores views through an anisotropic visibility field.
This avoids per-candidate rendering, but it captures parameter
uncertainty only indirectly. None of these methods addresses a
further inefficiency. Dense pools contain many near-duplicate views,
so much of the candidate evaluation is spent telling apart
observations that are effectively redundant.

A second line of work brings active 3DGS mapping onto robots.
RT-GuIDE~\cite{tao2024rtguiderealtimegaussiansplatting} couples
incremental Gaussian mapping with an information heuristic.
VISTA~\cite{nagami2026vista} adds semantics to plan task-relevant
trajectories. Splat-Nav~\cite{chen2025splat} plans safe corridors
inside Gaussian maps. Conflict-Aware Active
Perception~\cite{mollaei2026conflict} and Splat-CBF~\cite{khass2026splat} couples information acquisition
with a safety control barrier function. Closest to our anchor step,
Active next-best-view~\cite{khass2025active} optimizes expected
information gain over continuous camera poses on $\mathrm{SE}(3)$
under risk-aware path constraints. We build on this continuous
formulation, but we use the optimum to guide selection from a
discrete pool rather than as the target itself.

Our framework sits between these two groups. It keeps the
model-aware $\mathrm{SE}(3)$ guidance of parameter-space methods, but
it renders no candidate. Candidates are scored by an anchor leverage
score and pruned by fast subset selection. The result has the low
cost of geometric criteria while staying tied to an
information-theoretic optimum.

\section{Preliminaries}
\label{sec:Pre}

\begin{figure*}[t]
\centering
\includegraphics[width=\textwidth,trim={1.6cm 6.8cm 1.8cm 4.9cm},clip]{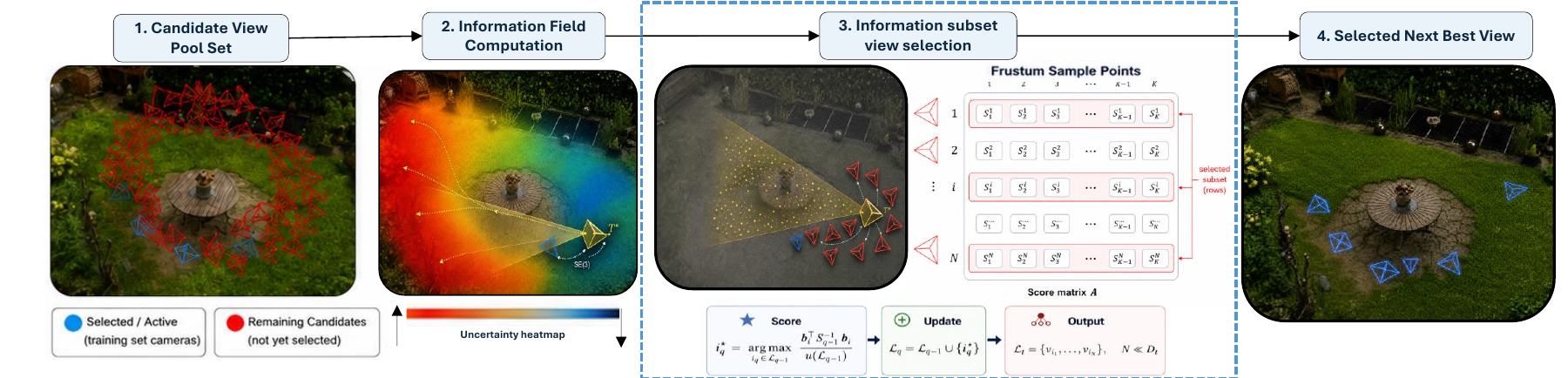}
\caption{\textbf{System overview.} Given a candidate view pool
$\mathcal{R}_t$ (red) around the active views $\mathcal{A}_t$
(blue), a virtual anchor is optimized to locate
a maximally informative region of pose space (uncertainty heatmap).
Candidates are scored against the anchor's frustum sample geometry
to form a score matrix, from which fast greedy ridge-leverage
selection extracts a compact, non-redundant subset views.}
\label{fig:system_overview}
\vspace{-0.4cm}
\end{figure*}

\subsection{3D Gaussian Splatting}
\label{sec:prelim-3dgs}
3D Gaussian Splatting~\cite{kerbl20233d} represents a scene as an
explicit set of anisotropic primitives,
\begin{equation}
  \mathcal{G} = \{g_i\}_{i=1}^{N}, \qquad
  g_i = \bigl(\boldsymbol{\mu}_i, \mathbf{R}_i, \mathbf{s}_i,
              o_i, \mathbf{c}_i\bigr),
\end{equation}
with center $\boldsymbol{\mu}_i \in \mathbb{R}^3$, rotation
$\mathbf{R}_i \in \mathrm{SO}(3)$, and scale
$\mathbf{s}_i \in \mathbb{R}^3_+$, which together define the
covariance $\boldsymbol{\Sigma}_i = \mathbf{R}_i
\operatorname{diag}(\mathbf{s}_i^2)\mathbf{R}_i^{\top}$, opacity
$o_i$, and spherical-harmonic coefficients $\mathbf{c}_i$ encoding
view-dependent appearance. For a view $v$, the visible primitives
are projected, depth-ordered, and alpha-composited at each pixel
$u$,
\begin{equation}
  f_v(u;\boldsymbol{\theta}) = \sum_{i \in \mathcal{N}_v(u)}
    \tau_{v,i}(u)\,\alpha_{v,i}(u)\,
    \mathbf{c}_i\bigl(\mathbf{d}_{v,i}\bigr),
\end{equation}
where $\boldsymbol{\theta}$ collects the Gaussian parameters and
$\mathcal{N}_v(u)$ is the ordered set of primitives contributing to
$u$. Given the acquired views $\mathcal{A}_t$ with reference images
$\{f_v^{\mathrm{ref}}\}_{v \in \mathcal{A}_t}$, the parameters are
estimated by minimizing a photometric objective,
\begin{equation}
  \hat{\boldsymbol{\theta}}_t = \arg\min_{\boldsymbol{\theta}}
    \sum_{v \in \mathcal{A}_t}
    \mathcal{L}_{\mathrm{photo}}\!\bigl(
      f_v^{\mathrm{ref}}, f_v(\boldsymbol{\theta})\bigr).
\end{equation}
Rasterization is differentiable, so each view contributes gradients
only to the subset of primitives it observes~\cite{matsuki2024gaussian}.


\subsection{Expected Information Gain for View Selection}
\label{sec:prelim-eig}

Active acquisition selects the view whose observation is expected to
most reduce uncertainty in the estimated scene
parameters~\cite{houlsby2011bayesian,kirsch2022unifying}. Let $Y_v$
denote the image predicted at candidate $v$ through
$f_v(\boldsymbol{\theta})$, and let $\hat{\boldsymbol{\theta}}_t$ be
the current estimate. The utility of $v$ is the mutual information
between the map parameters and $Y_v$, which equals the expected
reduction in posterior entropy $H[\cdot]$ after acquiring
$Y_v$~\cite{kirsch2022unifying},
\begin{equation}
  \mathcal{I}(\boldsymbol{\theta}; Y_v)
    = H[\boldsymbol{\theta}] - H[\boldsymbol{\theta} \mid Y_v].
\end{equation}
The observed information at view $v$ is the negative Hessian of the
rendering log-likelihood evaluated at $\hat{\boldsymbol{\theta}}_t$,
\begin{equation}
  \mathbf{H}_v = -\,\nabla_{\boldsymbol{\theta}}^{2}
    \log p(Y_v \mid \boldsymbol{\theta})
    \Big|_{\boldsymbol{\theta}=\hat{\boldsymbol{\theta}}_t},
  \label{eq:hessian-def}
\end{equation}
and the accumulated information over the acquired set is
$\hat{\mathbf{H}}_t := \sum_{v \in \mathcal{A}_t} \mathbf{H}_v +
\lambda \mathbf{I} \succ \mathbf{0}$. Under the Laplace
approximation, the entropy reduction admits the closed
form~\cite{jiang2024fisherrf,khass2025active}
\begin{equation}
\log\det \ \!\bigl(
    \mathbf{I} + \mathbf{H}_v \hat{\mathbf{H}}_t^{-1}\bigr),
  \label{eq:logdet-eig}
\end{equation}
where $\hat{\mathbf{H}}_t^{-1}$ encodes the uncertainty remaining in the
current map and $\mathbf{H}_v$ quantifies how strongly the candidate
constrains it. For the differentiable rendering model
$f_v(\boldsymbol{\theta})$, the observed information reduces to its
Gauss--Newton form, and the diagonal
approximation~\cite{jiang2024fisherrf,khass2025active} yields
\begin{equation}
  \mathbf{H}_v \approx
    \operatorname{diag}\!\bigl(\mathbf{J}_v^{\top}\mathbf{J}_v\bigr),
  \qquad
  \mathbf{J}_v := \nabla_{\boldsymbol{\theta}}\,
    f_v(\boldsymbol{\theta})\big|_{\hat{\boldsymbol{\theta}}_t}.
  \label{eq:diag-fisher}
\end{equation}
The next-best view maximizes the EIG over the candidate pool
$\mathcal{R}_t$ with cardinality $D_t = |\mathcal{R}_t|$,
\begin{equation}
  v_t^{\text{best}} = \arg\max_{v \in \mathcal{R}_t}\ \log\det \ \!\bigl(
    \mathbf{I} + \mathbf{H}_v \hat{\mathbf{H}}_t^{-1}\bigr).
  \label{eq:nbv-rule}
\end{equation}
Approximating Eq.~\eqref{eq:logdet-eig} recovers the
trace surrogate $\operatorname{tr}\!\left(\mathbf{H}_v\,\hat{\mathbf{H}}_t^{-1}\right) \ge \log\det\ \!\bigl(\mathbf{I}+\mathbf{H}_v\hat{\mathbf{H}}_t^{-1}\bigr)$~\cite{kirsch2022unifying},
which we use as a smooth objective for continuous anchor optimization.

\section{Efficient Next Best View Selection}
\label{sec:gradient_diverse_proposals}

Direct maximization of Eq.~\eqref{eq:nbv-rule} over the full candidate pool
$\mathcal{R}_t$ is accurate but expensive; each evaluation of
Eq.~\eqref{eq:logdet-eig} requires a differentiable render--backward
pass over the complete camera pool, and this cost grows with both the model size and the pool size
$D_t=\lvert\mathcal{R}_t\rvert$. Densely sampled image collections further contain many cameras with strongly overlapping fields of view, so a large fraction of this computation is spent distinguishing nearly redundant observations. The proposed framework addresses both limitations by decoupling the search for an informative region of pose space from the selection of executable cameras. The information gain objective is maximized over $\mathrm{SE}(3)$, and its local observation geometry is then used
to extract a compact set of relevant, non-redundant candidates. This reduces the effective search from $D_t$ cameras to $N\ll D_t$ proposals while retaining the
views most consistent with the continuous information optimum.


\subsection{Virtual Anchor Optimization }
\label{sec:continuous_information_optimization}
At selection event $t$, let $\mathcal{R}_t=\{T_v\}_{v=1}^{D_t}$ denote the
remaining candidate poses and let $\hat{\mathbf{H}}_t\succ\mathbf{0}$ denote the accumulated information from the active train set $\mathcal{A}_t$. Rather than scoring each discrete candidate,
 a virtual pose $T_t^{\star}
\in \mathrm{SE}(3)$ is optimized directly on the pose manifold to
mark a locally informative region.

For a view pose $T$, let ${\mathbf{H}}_T\succeq\mathbf{0}$ denote the Fisher information
contributed by its predicted observation. Under the Laplace approximation, we adopt the expected information gain of pose $T$ from Eq.~\eqref{eq:logdet-eig} by the first-order trace surrogate
\begin{equation}
  \psi_t(T) \;:=\; \operatorname{tr}\ \!\bigl(
    {\mathbf{H}}_T \hat{\mathbf{H}}_t^{-1}\bigr).
  \label{eq:trace-surrogate}
\end{equation}
Since the diagonal
Fisher term is differentiable in the rendering parameters~\cite{matsuki2024gaussian,khass2025active}, and the rendering is
differentiable in the camera pose, $\psi_t(T)$ is differentiable with
respect to $T$ and can be optimized directly over the pose manifold.
The continuous information anchor is, 
\begin{equation}
    T_t^{\star}
    =
    \arg\max_{T\in\mathrm{SE}(3)} 
    \psi_t(T)
    \label{eq:continuous_nbv_anchor}
\end{equation}
which steers the virtual camera toward observations expected to resolve the
current model uncertainty. Eq.~\eqref{eq:continuous_nbv_anchor} is solved by the Riemannian gradient
ascent method~\cite{khass2025active}. At iteration $k$, the Euclidean gradient of
$\psi_t$ is mapped to a twist
$\boldsymbol{\xi}_k\in\mathfrak{se}(3)\cong\mathbb{R}^{6}$ in the local tangent
space at $T_k$, and the pose is updated through the exponential map retraction,
\begin{equation}
    T^{k+1}
    = T^k \,\exp\!\bigl(\eta\,\boldsymbol{\xi}_k^{\wedge}\bigr),
    \qquad
    \boldsymbol{\xi}_k
    = \nabla_{T}\,\psi_t(T)\big|_{T^{k}},
    \label{eq:se3_update}
\end{equation}
with step size $\eta > 0$,  $(\cdot)^{\wedge}: \mathbb{R}^6 \to
\mathfrak{se}(3)$ the twist hat map, and $\exp:\mathfrak{se}(3) \to \mathrm{SE}(3)$ the matrix exponential. The retraction keeps every iterate on the manifold and preserves the coupling between
translational and rotational components.
The resulting $T_t^\star$ is an information anchor, not the  selected view. In general it might not coincide with a camera in $\mathcal{R}_t$, which motivates a principled association between the continuous optimum and the discrete candidate pool.

\subsection{Candidate View Scoring}
\label{sec:pool_view_representation}

Associating the anchor $T_t^{\star}$ with its nearest camera is inadequate; pose distance is a property of the manifold alone and carries no information about the scene content a camera observes. Two candidates equidistant from $T_t^{\star}$ may see disjoint
regions of the map, while distinct poses may induce nearly identical observations. Each candidate is therefore represented not by its pose, but by how faithfully it reproduces the viewing geometry of the information anchor.

The anchor viewing volume is discretized into $K$ frustum support
samples,
\begin{equation}
    \mathcal{X}_t = \{\mathbf{x}_k\}_{k=1}^{K},
    \qquad
    \mathbf{x}_k\in\mathcal{F}(T_t^{\star}),
    \label{eq:support_samples}
\end{equation}
where $\mathcal{F}(T_t^{\star})$ denotes the viewing frustum of the anchor pose.
These samples act as a shared spatial basis against which every remaining camera
is evaluated.
For a candidate $v\in\mathcal{R}_t$ with pose $T_v=(R_v,\mathbf{p}_v)$ and optical
axis $\mathbf{d}_v$, sample
$\mathbf{x}_k$ is frustum contained when $\mathbf{x}_k \in \mathcal{F}(T_v)$, and the corresponding viewing
direction from the camera center is
\begin{equation}
    \chi_{vk}
    = \mathbf{1}_{\mathcal{F}(T_v)}(\mathbf{x}_k),
    \qquad
    \hat{\mathbf{r}}_{vk}
    = \frac{\mathbf{x}_k-\mathbf{p}_v}
           {\lVert\mathbf{x}_k-\mathbf{p}_v\rVert_2}.
    \label{eq:frustum-visibility}
\end{equation}
where $\chi_{vk}\in\{0,1\}$ marks whether candidate $v$ observes
$\mathbf{x}_k$, and $\hat{\mathbf{r}}_{vk}$ is the viewing direction.


The global consistency of candidate $v$ with the anchor is summarized by a
scalar weight $\rho_v\in[0,1]$, defined as the product of positional,
orientational, and coverage factors,
\begin{equation}
    \rho_v
    = \underbrace{\exp\!\left(-\frac{\lVert\mathbf{p}_v-\mathbf{p}_t^\star\rVert_2^{2}}
        {2\sigma_p^{2}}\right)}_{\text{position}}
      \;\cdot\;
      \underbrace{\bigl[\mathbf{d}_v^\top\mathbf{d}_t^\star\bigr]_{+}^{2}}_{\text{orientation}}
      \;\cdot\;
      \underbrace{\frac{1}{K}\sum_{k=1}^{K}\chi_{vk}}_{\text{coverage}},
    \label{eq:anchor_pose_consistency}
\end{equation}
with $\sigma_p$ a positional tolerance and $[\,\cdot\,]_{+}=\max(\cdot,0)$.
The orientation factor $[\mathbf{d}_v^\top\mathbf{d}_t^\star]_{+}^{2}$ assigns zero weight to back facing cameras and grows quadratically as orientations align. 

Combining the global consistency weight $\rho_v$ with per-sample geometry yields the anchor coverage matrix
$\mathbf{M_t}\in\mathbb{R}^{D_t\times K}$, whose entries are
\begin{equation}
    [\mathbf{M}_t]_{vk}
    = \rho_v^{\gamma}\,\underbrace{\chi_{vk}\,
      \frac{\bigl[\mathbf{d}_v^\top\hat{\mathbf{r}}_{vk}\bigr]_{+}}
           {\lVert\mathbf{x}_k-\mathbf{p}_v\rVert_2+\epsilon}}_{\displaystyle\kappa_{vk}},
    \label{eq:anchor_observation_matrix}
\end{equation}
where $\gamma>0$ controls the sharpness of anchor weighting and
$\epsilon > 0$ is a small numerical regularizer. The per-sample observation kernel $\kappa_{vk}$ is nonzero only for visible samples ($\chi_{vk}=1$), rewards well centered viewing directions through the rectified alignment $[\mathbf{d}_v^\top\hat{\mathbf{r}}_{vk}]_{+}$, and downweights distant or oblique baselines through the inverse range. 
The row weight $\rho_v^{\gamma}$ scales the entire profile of candidate $v$, suppressing cameras that do not preserve the
position, orientation, or spatial support of the anchor.

Each row of $\mathbf{M}_t$ is the observation profile of one candidate over the $\mathcal{X}_t$, and each column describes how the pool candidate views observe
a single location $\mathbf{x}_k$. 
The entry $[\mathbf{M}_t]_{vk}$ therefore measures how effectively candidate view $v$ observes the anchor viewing volume, jointly accounting for visibility, viewing direction, range, and anchor consistency.
Candidates with similar rows induce redundant observation geometry, whereas dissimilar rows provide complementary coverage of the anchor volume.

\subsection{Fast Subset Selection via Ridge Leverage}
\label{sec:diverse_proposal_selection}

Selecting a compact proposal set from the anchor coverage matrix
$\mathbf{M}_t\in\mathbb{R}^{D_t\times K}$ requires comparing candidate profiles
that live in a $K$-dimensional support space. Since the number of frustum support
samples $K$ can be large, the rows of $\mathbf{M}$ are first embedded into a
$d$-dimensional space through a fixed Johnson--Lindenstrauss
projection~\cite{johnson1984extensions,nelson2020dimensionality},
\begin{equation}
    \mathbf{B}_t = \mathbf{M}_t\,\boldsymbol{\Pi},
    \qquad
    \boldsymbol{\Pi}\in\mathbb{R}^{K\times d},
    \qquad
    d\ll K.
    \label{eq:jl_camera_projection}
\end{equation}

Let $\mathbf{b}_v \in \mathbb{R}^d$ denote the row
of $\mathbf{B}_t$ associated with candidate $v \in \mathcal{R}_t$.
Ranking candidates by profile magnitude $\lVert\mathbf{b}_v\rVert_2$
alone can return several cameras that all encode the same dominant observation direction; each is informative in isolation, yet together they leave the remaining directions of the anchor volume
unconstrained. 
\begin{algorithm}[t]
\caption{\textsc{AGILE-GS}: Anchor-Guided Fast Next-Best-View Selection for Active 3D Gaussian Splatting}
\label{alg:diverse_nbv}
\begin{algorithmic}[1]
\State \textbf{Input:} Active views $\mathcal{A}_t$, candidate pool $\mathcal{R}_t$, accumulated information $\hat{\mathbf{H}}_t$, proposal budget $N$
\State \textbf{Output:} Next-best view $v_t^\star$ and proposal set $\mathcal{L}_t$
\State Optimize the continuous information anchor on $\mathrm{SE}(3)$:
\[
T_t^{\star}
    \leftarrow
    \arg\max_{T\in\mathrm{SE}(3)} 
    \psi_t(T)
\]
\State Build the coverage matrix $\mathbf{M}_t$
\State Project the profiles to $\mathbf{B}_t \leftarrow \mathbf{M}_t\boldsymbol{\Pi}$,
with rows $\{\mathbf{b}_v\}$
\State Initialize $\mathcal{L}_0$ and
$\mathbf{S}_0^{-1} \leftarrow \lambda_G^{-1}\mathbf{I}_d$
\For{$q = 1,\ldots,\min(N,\,D_t)$}
    \State \textbf{Select} the highest conditional ridge-leverage candidate:
    $v_q^\star \leftarrow \arg\max_{v\notin\mathcal{L}_{q-1}}
    \mathbf{b}_v^\top\mathbf{S}_{q-1}^{-1}\mathbf{b}_v$
    \State \textbf{Append} $v_q^\star$ to the proposal set:
    $\mathcal{L}_q \leftarrow \mathcal{L}_{q-1}\cup\{v_q^\star\}$
    \State \textbf{Update} $\mathbf{S}_q^{-1}$ by a rank-one
    Sherman--Morrison
\EndFor
\State 
$\mathcal{L}_t \leftarrow \{v_1^\star,\ldots,v_N^\star\}$
\State \textbf{return} leading proposal $v_t^\star \leftarrow v_1^\star$ and
$\mathcal{L}_t$
\end{algorithmic}
\end{algorithm}
The proposal set must therefore reward informativeness \emph{and} penalize overlap with
already selected candidate views. Therefore we employ the candidate view set selection objective as 
\begin{equation}
\begin{aligned}
    &\mathcal{L}_t^\star
    = \arg\max_{\mathcal{L}\subseteq\mathcal{R}_t,\,\lvert\mathcal{L}\rvert\le N}
      F_t(\mathcal{L}),
     \end{aligned}
\end{equation}
     \begin{equation}
\begin{aligned}
    &F_t(\mathcal{L})
    = \log\det\!\Bigl(\lambda_G\mathbf{I}_d
      + \!\sum_{v\in\mathcal{L}}\!\mathbf{b}_v\mathbf{b}_v^\top\Bigr)
      - d\log\lambda_G,
    \label{eq:camera_doptimal_objective}
\end{aligned}
\end{equation}
where $\lambda_G>0$. Hence, a candidate aligned with a well represented direction
expands volume $F_t(\mathcal{L})$ little, whereas one contributing a new direction expands it
substantially. Maximizing $F_t$ thus selects the most informative cameras
subject to mutual complementarity and candidate diversity.

Let $\mathcal{L}_{q-1}$ be the first $q-1$ selections and
\begin{equation}
    \mathbf{S}_{q-1}
    = \lambda_G\mathbf{I}_d
      + \sum_{v\in\mathcal{L}_{q-1}}\mathbf{b}_v\mathbf{b}_v^\top.
    \label{eq:geometric_information_matrix}
\end{equation}
By the matrix determinant lemma, the marginal gain of adding candidate
$v$ is
\begin{equation}
F_t(\mathcal{L}_{q-1}\cup\{v\}) - F_t(\mathcal{L}_{q-1})
    = \log\ \!\bigl(1+\mathbf{b}_v^\top\mathbf{S}_{q-1}^{-1}\mathbf{b}_v\bigr).
    \label{eq:camera_marginal_gain}
\end{equation}
Since $\log(1+x)$ is strictly increasing, the greedy step reduces to maximizing the
conditional quadratic score
\begin{equation}
    v_q^\star
    = \arg\max_{v\notin\mathcal{L}_{q-1}}
      \underbrace{\mathbf{b}_v^\top\mathbf{S}_{q-1}^{-1}\mathbf{b}_v}_{\ell_v(\mathcal{L}_{q-1})},
    \qquad
    \mathcal{L}_q = \mathcal{L}_{q-1}\cup\{v_q^\star\},
    \label{eq:greedy_camera_leverage}
\end{equation}
where $\ell_v(\mathcal{L}_{q-1})$ is a conditional ridge-leverage
score~\cite{cohen2016online}. The inverse is maintained without needing
recomputation in each step through the Sherman--Morrison identity,
\begin{equation}
    \mathbf{S}_{q}^{-1}
    = \mathbf{S}_{q-1}^{-1}
      - \frac{\mathbf{z}_q\mathbf{z}_q^\top}
             {1+\mathbf{b}_{v_q^\star}^{\top}\mathbf{z}_q},
    \qquad
    \mathbf{z}_q = \mathbf{S}_{q-1}^{-1}\mathbf{b}_{v_q^\star}.
    \label{eq:greedy_inverse_update}
\end{equation}

Eq.~\eqref{eq:greedy_camera_leverage} explains the redundancy suppression
directly. At the first iteration
$\mathbf{S}_0=\lambda_G\mathbf{I}_d$ and
$\ell_v=\lVert\mathbf{b}_v\rVert_2^{2}/\lambda_G$, so the leading proposal is the candidate with the strongest anchor observation profile. 
Each subsequent selection contracts $\mathbf{S}_q^{-1}$
along the chosen direction, so the score of any later candidate is attenuated in proportion to its alignment with the already selected subspace; a camera that is nearly collinear with a chosen one collapses to a small score regardless of its raw magnitude. Each greedy step selects the candidate with the highest conditional score, which rewards a view only when it is both informative and complementary to the current selection.
Redundant candidates are passed over but remain in $\mathcal{R}_t$ and stay eligible at later events.

After $N$ iterations, the ordered proposal set is
\begin{equation}
    \mathcal{L}_t = \bigl\{v_1^\star,\dots,v_N^\star\bigr\},
    \qquad N\ll D_t.
    \label{eq:diverse_candidate_set}
\end{equation}
The AGILE-GS algorithm returns the leading proposal $v_1^\star$ directly as the next best view, since it maximizes the marginal gain with respect to the regularized feature space at zero per-candidate Fisher cost. With the accuracy oriented strategy, AGILE-GS+ algorithm instead reranks the full shortlist $\mathcal{L}_t$ by expected information gain (Sec.~\ref{sec:optimal}).

\section{model aware Reranking}
\label{sec:optimal}

The \textbf{AGILE-GS} algorithm~(\ref{alg:diverse_nbv}) in Sec.~\ref{sec:diverse_proposal_selection}
maximizes the marginal gain of the geometric objective in
Eq.~\eqref{eq:camera_doptimal_objective}.
This objective operates in the projected observation space and quantifies how faithfully a candidate preserves and complements the anchor's viewing geometry, but does not measure its effect on the uncertainty of individual 3DGS parameters. Two views with comparable geometric scores may constrain disjoint subsets of Gaussians and induce markedly different reductions in posterior uncertainty.

\textsc{AGILE-GS+} addresses this by reranking the compact proposal set by expected information gain evaluation. Let 
\begin{equation}
    \mathcal{L}_t = \{v_1^\star,\dots,v_N^\star\}\subseteq\mathcal{R}_t,
    \qquad N\ll D_t,
    \label{eq:proposal_set_recall}
\end{equation}
denote the ordered proposal set returned by algorithm~(\ref{alg:diverse_nbv}).
By construction, $\mathcal{L}_t$ is aligned with the continuous anchor and spans distinct directions of its observation space, so the model-space evaluation is confined to a small set of geometrically informative, non-redundant candidates rather than the full pool.

We quantify the utility of a candidate view $v$ by the reduction in posterior entropy introduced in Sec.~\ref{sec:prelim-eig}, \begin{equation} 
\phi_t(v) := \log\det\!\left( \mathbf{I} + \mathbf{H}_v\hat{\mathbf{H}}_t^{-1} \right),
\label{eq:plus_gain}
\end{equation}
where 
\( \hat{\mathbf{H}}_t = \sum_{\ell\in\mathcal{A}_t}\mathbf{H}_{\ell} +\lambda\mathbf{I}\succ0 \) 
denotes the information accumulated from the acquired views. Hence, $\phi_t(v)$ measures the information contributed by $v$ relative to parameter space that are already constrained by $\mathcal{A}_t$.
Since, under the diagonal Gauss–Newton approximation, both
$\mathbf{H}_v\approx\operatorname{diag}(\mathbf{J}_v^{\top}\mathbf{J}_v)$ and  $\hat{\mathbf{H}}_t$
are diagonal, the determinant in Eq.~\eqref{eq:plus_gain} factorizes over its entries and
\begin{equation}
\begin{aligned}
\phi_t(v) = \log\det\!\left( \mathbf{I} + \mathbf{H}_v\hat{\mathbf{H}}_t^{-1} \right)
&\approx
\log\!\prod_q \left( 1+ \frac{[\mathbf{H}_v]_{qq}} {[\hat{\mathbf{H}}_t]_{qq}} \right)
\\
&= \sum_{q}\log\!\left(1+\frac{[\mathbf{H}_v]_{qq}}{[\hat{\mathbf{H}}_t]_{qq}}\right),
\label{eq:plus_diagonal}
\end{aligned}
\end{equation}
where $q$ indexes the Gaussian parameters and $[\,\cdot\,]_{qq}$ denotes the
$q$-th diagonal entry. 

AGILE-GS+ algorithm selects the maximizer of Eq.~\eqref{eq:plus_diagonal} over the proposal set,
\begin{equation}
  v_t^{\star\star}=\arg\max_{v\in\mathcal{L}_t}\phi_t(v),\quad
  \hat {\mathbf{H}}_{t+1}=\hat{\mathbf{H}}_t+\mathbf{H}_{v_t^\star}
  \label{eq:plus_nbv}
\end{equation}
after which the view sets and accumulated information are updated:
\begin{equation}
  \mathcal{A}_{t+1}=\mathcal{A}_t\cup\{v_t^{\star\star}\},\quad
  \mathcal{R}_{t+1}=\mathcal{R}_t\setminus\{v_t^{\star\star}\}.
  \label{eq:plus_update}
\end{equation}
The two methods therefore differ only in how they exploit $\mathcal{L}_t$ set. AGILE-GS returns the leading proposal $v_1^\star$, whereas AGILE-GS+ performs $N$ Fisher evaluations, one per proposal.
\begin{algorithm}[t]
\caption{AGILE-GS+:Model Aware NBV Selection on Anchor Shortlist}
\label{alg:finegs_plus}
\begin{algorithmic}[1]
\State \textbf{Input:}$\mathcal{A}_t$, $\mathcal{R}_t$,
accumulated information $\hat{\mathbf{H}}_t$, budget $N$
\State \textbf{Output:} Next-best view $v_t^{\star\star}$
\State $\mathcal{L}_t\leftarrow\textsc{AGILE-GS}\bigl(\mathcal{A}_t,\mathcal{R}_t,\hat{\mathbf{H}}_t,N\bigr)$
\Comment{Algorithm.~\ref{alg:diverse_nbv};}
\ForAll{$v\in\mathcal{L}_t$}
    \State $\mathbf{H}_v\leftarrow\operatorname{diag}(\mathbf{J}_v^{\top}\mathbf{J}_v)$
    \State $\displaystyle\phi_t(v)\leftarrow\sum_{q}
           \log\bigl(1+[\mathbf{H}_v]_{qq}/[\hat{\mathbf{H}}_t]_{qq}\bigr)$
\EndFor
\State $v_t^{\star\star}\leftarrow\arg\max_{v\in\mathcal{L}_t}\phi_t(v)$
\State $\mathcal{A}_{t+1}\leftarrow\mathcal{A}_t\cup\{v_t^\star\}$,\;
       $\mathcal{R}_{t+1}\leftarrow\mathcal{R}_t\setminus\{v_t^{\star\star}\}$,\;
       \\ $\hat {\mathbf{H}}_{t+1}\leftarrow\hat{\mathbf{H}}_t+\mathbf{H}_{v_t^{\star\star}}$
\State \Return $v_t^{\star\star}$
\end{algorithmic}
\end{algorithm}
\vspace{-0.2 cm}

\begin{figure*}[t]
\centering
\includegraphics[
        width=\textwidth,
        trim={4.9cm 5.0cm 5.5cm 4.5cm},
        clip
    ]{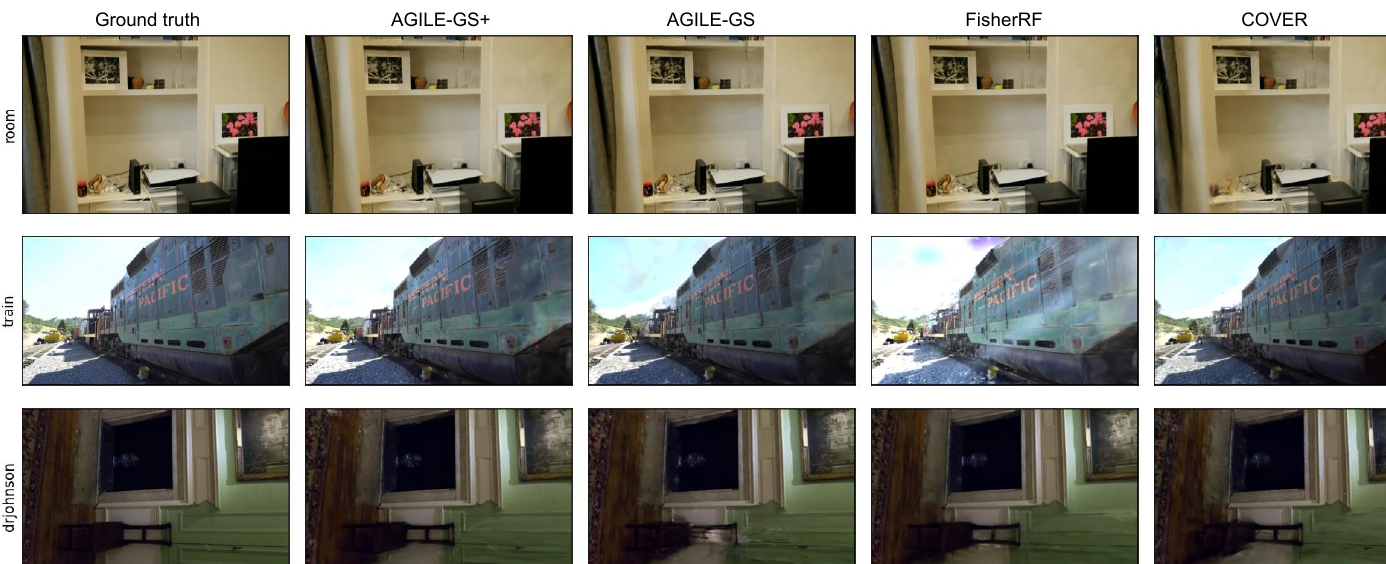}
\caption{
Qualitative evaluation renders of the 3DGS comparison on representative views from different view selection methods. 
}
\label{fig:qualitative}
\end{figure*}
\section{Experiments}
\label{sec:experiments}
We evaluate our proposed algorithms on progressive next-best-view
selection for 3DGS. We first compare reconstruction quality and selection efficiency against existing NBV baseline methods, then evaluate the methods under an embodied acquisition protocol, and finally ablate the candidate-selection strategy and the shortlist view pool size.

\subsection{Experimental Setup}
\label{sec:experimental_setup}

\paragraph{Datasets.}
We evaluate on three standard reconstruction benchmarks:
Mip-NeRF~360~\cite{barron2022mipnerf360}, Tanks and
Temples~\cite{Knapitsch2017}, and Deep
Blending~\cite{DeepBlending2018}, which cover diverse indoor and outdoor scenes, object scales, camera distributions, and geometric complexity. We additionally report embodied acquisition on our custom scenes. All experiments were run on a workstation with an NVIDIA RTX A2000 GPU, an Intel Core i9-13900K CPU.

\textbf{Baselines.}
We compare against three baselines: {Random} uniformly samples, {FisherRF}~\cite{jiang2024fisherrf} scores every candidate by information gain; and {COVER}~\cite{chen2026coverage} approximates Fisher information through geometric view coverage, favoring viewpoints that observe under-constrained regions. 
Our two variants differ in how $\mathcal{L}_t$ is used; {AGILE-GS} prioritizes selection efficiency by returning the leading ridge-leverage candidate to select an informative, non-redundant view without exhaustive Fisher evaluation, while {AGILE-GS+} evaluates expected information gain only over the
compact shortlist $\mathcal{L}_t$, retaining the accuracy of
Fisher-based selection while avoiding its cost over the full pool.

\textbf{Metrics.}
Reconstruction quality is measured with PSNR ($\uparrow$),
SSIM ($\uparrow$), and LPIPS ($\downarrow$). Since the shared
3DGS optimization is identical across methods, the computational
overhead of view selection is isolated by reporting the selection
latency per acquisition event (s/event).

\textbf{Setup.}
Each scene is initialized with 10 training views. During progressive training, every 200 gradient steps a view is selected from the remaining candidate pool $\mathcal{R}_t$ and added to the active training set $\mathcal{A}_t$.
All methods are trained for $30\text{K}$ gradient steps
under an identical acquisition budget and reconstruction
configuration, so the only computational difference lies in the NBV decision itself.

\subsection{Results}
\label{sec:results}

Table~\ref{tab:main_results} reports reconstruction quality and
average selection latency across the three benchmarks. AGILE-GS+ achieves the best or second-best average reconstruction metrics, while the efficiency-oriented AGILE-GS also maintains strong reconstruction quality despite avoiding candidate-level Fisher reranking.
The representative reconstructions in Fig.~\ref{fig:qualitative}
are consistent with these quantitative results.

\begin{figure}[t]
\centering
\includegraphics[width=\linewidth,
        trim={5.7cm 4.8cm 7.8cm 3.5cm},
        clip]{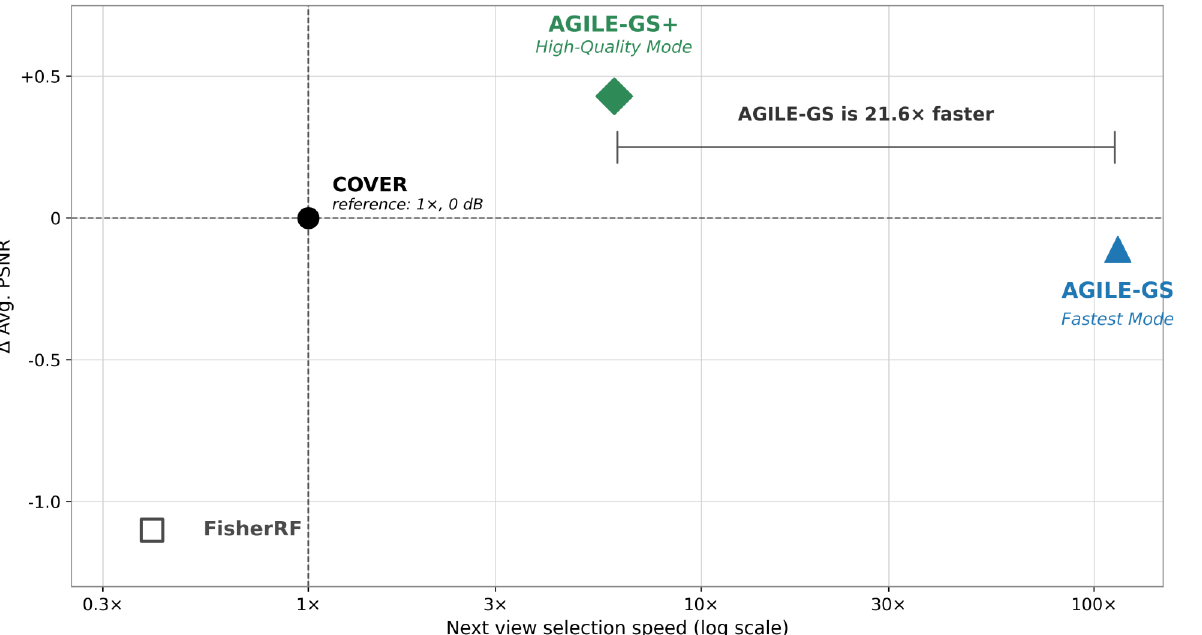}
\caption{
Accuracy--efficiency trade-off relative to COVER, normalized to
$1\times$ selection speed and $0$\,dB PSNR difference. 
}
\label{fig:tradeoff}
\end{figure}

The efficiency gap is substantially larger than the remaining
differences in reconstruction accuracy. AGILE-GS is approximately
$126\times$ faster than COVER and $379\times$ faster than FisherRF per selection event, while AGILE-GS+ is $5.8\times$ and $17.6\times$ faster respectively and still attains higher average PSNR than both. 
Fig.~\ref{fig:tradeoff} makes this trade-off explicit by
normalizing accuracy and selection speed to COVER.
AGILE-GS+ lies above and to the right of the reference, improving PSNR while increasing selection speed,
and AGILE-GS extends a further
$21.6\times$ speedup at only a small PSNR cost.
Together, AGILE-GS and AGILE-GS+ provide two complementary operating modes. AGILE-GS for maximum selection speed and AGILE-GS+ for higher reconstruction quality while remaining substantially faster than existing baselines.\footnote{COVER~\cite{chen2026coverage} originally reports $3.5$\,s
per selection and $23.9$\,s for FisherRF on undisclosed hardware, which may be more capable than ours. Still, AGILE-GS+ is $2.2\times$ and AGILE-GS is approximately $47\times$ faster than COVER.} 
Random is exceeding Fisher\-RF on Mip-NeRF 360; the same ordering is reported by
COVER, since human captured benchmark are already well distributed and uniform sampling inherits that feature~\cite{chen2026coverage}. 

We further evaluate the methods under the embodied acquisition
protocol of COVER~\cite{chen2026coverage}, in which each acquisition is restricted to the $K{=}5$ nearest candidate frames around the current view rather than the full pool. As shown in Table~\ref{tab:embodied}, AGILE-GS+ remains competitive with COVER under these motion constraints and outperforms Random and FisherRF on both the Mip-NeRF~360 and custom evaluations.

\begin{table}[t]
    \centering
    \setlength{\tabcolsep}{3.2pt}
    \renewcommand{\arraystretch}{1.08}
    \resizebox{\columnwidth}{!}{%
    \begin{tabular}{lccc ccc}
        \toprule
        \multirow{2}{*}{\textbf{Embodied Method}}
        & \multicolumn{3}{c}{\textbf{Mip-NeRF 360}}
        & \multicolumn{3}{c}{\textbf{Custom}} \\
        \cmidrule(lr){2-4}\cmidrule(lr){5-7}
        & PSNR$\uparrow$ & SSIM$\uparrow$ & LPIPS$\downarrow$
        & PSNR$\uparrow$ & SSIM$\uparrow$ & LPIPS$\downarrow$ \\
        \midrule
        Random
        & 24.786 & 0.6712 & 0.2071
        & 20.186 & 0.5441 & 0.2633 \\

        FisherRF
        & 24.672 & 0.6712 & 0.2083
        & 20.400 & 0.5550 & 0.2512 \\

        COVER
        & \textbf{26.571} & \textbf{0.8158} & \textbf{0.1838}
        & \underline{22.629} & \underline{0.6912} & \underline{0.2186} \\
        \midrule
        AGILE-GS
        & 26.134 & 0.7960 & \underline{0.1882}
        & 22.373 & 0.6514 & 0.2385 \\

        AGILE-GS+
        & \underline{26.431} & \underline{0.8073} & {0.1895}
        & \textbf{23.835} &\textbf{0.7657} & \textbf{0.1962} \\
        \bottomrule
    \end{tabular}}
        \caption{
Embodied view selection restricted to the K=5 nearest candidate frames.
}
\label{tab:embodied}
\end{table}

\begin{figure}[t]
    \centering
    \includegraphics[
        width=0.50\textwidth,height=0.17\textheight,
        trim={6.0cm 8.7cm 6.1cm 5.5cm},
        clip
    ]{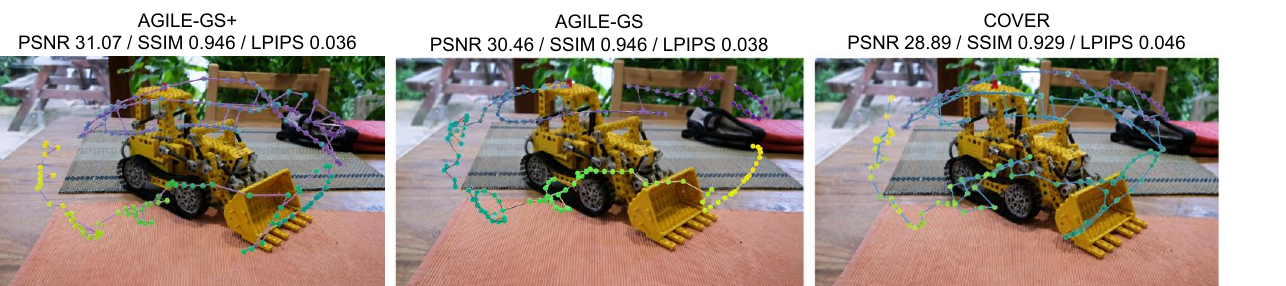}
\caption{ Selected views and reconstruction quality. The colored camera trajectories visualize the ordering of acquired views. } \label{fig:view_trajectory}
\end{figure}

Beyond reconstruction accuracy, Fig.~\ref{fig:view_trajectory}
exposes a structural difference in the acquisition trajectories.
Although the trajectories themselves differ considerably, the methods largely agree on which views are informative. Over the 150 selected views, AGILE-GS+ and AGILE-GS share $72.3\%$ and
$69.3\%$ of their selections with COVER, respectively. The principal difference is not the informative support recovered from the scene, but the order in which that support is acquired. Successive AGILE-GS selections remain more local in the camera sequence, reducing the mean frame-index transition by approximately $41\%$ relative to COVER. Next best view selection policy for a physical robot cannot be considered independently of the motion required between consecutive observations; widely separated NBV decisions introduce additional travel, trajectory planning, and execution cost even when the selected views are individually informative. Such continuity reduces the discrepancy between discrete NBV optimization and physically executable sensing, making the selected views better suited to continuous robotic acquisition.


\begin{table*}[t]
\centering
\setlength{\tabcolsep}{4.2pt}
\renewcommand{\arraystretch}{1.12}

\begin{minipage}[t]{0.77\textwidth}
\centering

\vspace{1.5mm}

\resizebox{\linewidth}{!}{
\begin{tabular}{
l
ccc!{\vrule width 0.4pt}
ccc!{\vrule width 0.4pt}
ccc
}
\toprule
\multirow{2}{*}{Method}
&
\multicolumn{3}{c}{Mip-NeRF 360}
&
\multicolumn{3}{c}{Tanks \& Temples}
&
\multicolumn{3}{c}{Deep Blending}
\\
\cmidrule(lr){2-4}
\cmidrule(lr){5-7}
\cmidrule(lr){8-10}
&
PSNR$\uparrow$ & SSIM$\uparrow$ & LPIPS$\downarrow$
&
PSNR$\uparrow$ & SSIM$\uparrow$ & LPIPS$\downarrow$
&
PSNR$\uparrow$ & SSIM$\uparrow$ & LPIPS$\downarrow$
\\
\midrule

Random 
& 26.977
& {0.8306}
& 0.1165
& 20.820
& 0.7693
& 0.1932
& 24.264
& 0.6245
& 0.2834
\\

FisherRF 
& 26.515
& 0.8225
& 0.1312
& 21.214
& 0.7816
& 0.1795
& 24.578
& 0.6856
& 0.2246
\\

COVER
& \underline{27.231}
& 0.8283
& \underline{0.1145}
& \best{22.161}
& \best{0.8028}
& \best{0.1587}
& 26.154
& 0.7196
& 0.1977
\\

\midrule

AGILE-GS
& 26.996
& \underline{0.8310}
& 0.1159
& 21.823
& 0.7872
& 0.1743
& \underline{26.412}
& \underline{0.8344}
& \underline{0.1704}
\\

AGILE-GS+
& \best{27.523}
& \best{0.8467}
& \best{0.1112}
& \underline{22.003}
& \underline{0.7986}
& \underline{0.1683}
& \best{27.328}
& \best{0.8450}
& \best{0.1632}
\\

\bottomrule
\end{tabular}
}
\end{minipage}
\hfill
\begin{minipage}[t]{0.214\textwidth}
\centering

\vspace{1.5mm}

\resizebox{\linewidth}{!}{
\begin{tabular}{
c!{\vrule width 0.4pt}c
}
\toprule
\multicolumn{1}{c}{\multirow{2}{*}{\shortstack{Avg.\\PSNR$\uparrow$}}}
&
\multicolumn{1}{c}{\multirow{2}{*}{\shortstack{Avg. Selection\\Time (s)$\downarrow$}}}
\\
&
\\
\midrule

24.020
& 0.002
\\

24.103
& 28.030
\\

\underline{25.182}
& 9.330
\\

\midrule

25.077
& \multicolumn{1}{>{\columncolor{gray!35}}c}{
    \rule{0pt}{2.6ex}0.074
}
\\

\best{25.618}
& \multicolumn{1}{>{\columncolor{gray!15}}c}{
    \rule{0pt}{2.6ex}1.596
}
\\

\bottomrule
\end{tabular}
}
\end{minipage}

\caption{
Reconstruction quality and next-view selection efficiency on
Mip-NeRF~360, Tanks~\&~Temples, and Deep Blending. Reconstruction
metrics are averaged over the evaluated scenes in each benchmark;
selection time denotes the average latency per acquisition
event. Best and second-best reconstruction results are shown in
\textbf{bold} and \underline{underlined}, respectively.
}
\label{tab:main_results}
\end{table*}


\subsection{Ablation Studies} \label{sec:ablation} 
Table~\ref{tab:candidate_ablation} compares AGILE-GS against three
alternative strategies for selecting the proposal set from
$\mathcal{R}_t$: {$K$-nearest} retains the $N$ candidates
closest to $T_t^{\star}$ under an SE(3) metric; {top
independent norm} keeps the $N$ rows of $\mathbf{B}_t$ with the
largest $\lVert\mathbf{b}_v\rVert_2$; and {online row
sampling} which is a streaming baseline that processes candidate view descriptors sequentially and makes an immediate keep-or-discard decision without access to the complete candidate pool. Specifically, it uses the online $\lambda$-ridge leverage score~\cite{cohen2016online},
\[
\ell_i = a_i^\top\!\left(A_{i-1}^\top A_{i-1}+\lambda I\right)^{-1}\!a_i,
\]
to retain views that contribute directions not already well represented by previously accepted samples. This makes it particularly attractive for robotic settings, where observations arrive online, and future viewpoints are not available in advance. Nevertheless, Table~\ref{tab:candidate_ablation} shows that AGILE-GS achieves stronger reconstruction quality by explicitly selecting a compact batch of informative and complementary views. The result indicates that the proposed conditional ridge-leverage objective not only reduces selection cost, but also suppresses redundant observations that contribute little additional information to the reconstructed map.

 Table~\ref{tab:budget_ablation} varies the proposal budget $N$ used by AGILE-GS+ algorithm. Reconstruction quality remains nearly unchanged over a broad range of shortlist sizes, while the selection cost increases as more candidates are passed to the Fisher reranking stage. A compact shortlist therefore captures most of the useful information required for accurate selection, validating the central design of AGILE-GS+. Expensive model aware evaluation is needed only over a small subset of the original candidate pool. We use $N=10$ as the default setting, providing a strong balance between reconstruction quality and selection efficiency.

\begin{table}[t]
\centering
\scriptsize
\setlength{\tabcolsep}{3.5pt}
\renewcommand{\arraystretch}{2.05}
\begin{tabular}{llcccc}
\toprule
Dataset & Method 
& PSNR$\uparrow$ 
& SSIM$\uparrow$ 
& LPIPS$\downarrow$ 
& Speed up \\
\midrule
\multirow{4}{*}{\shortstack{Mip-NeRF 360\\and Custom}}
&  $K$-nearest Views      & 26.545 & 0.8603 & 0.2081 & 1.52$\times$ \\
& Online Row Sampling   & 27.202 & 0.8840 &  0.1600 & 1.48$\times$ \\ 
& Top independent norm   & 26.883 & 0.8845 & 0.1639 & 1.12$\times$ \\
& AGILE-GS (Ours)             & 27.692 & 0.9155 & 0.1413 & 1$\times$  \\
\bottomrule
\end{tabular}
\caption{Batch view selection ablation averaged over Mip-NeRF~360 and the custom dataset.}
\label{tab:candidate_ablation}
\end{table}


\begin{table}[t]
    \centering
    \small
    \setlength{\tabcolsep}{5.5pt}
    \renewcommand{\arraystretch}{1.10}
    \begin{tabular}{@{}c ccc c@{}}
        \toprule
        \multirow{2}{*}{
        \shortstack{AGILE-GS+ \\ {Budget $N$}}} 
        & \multicolumn{3}{c}{{Avg Mip-NeRF 360--Custom}}
        & \multirow{2}{*}{\shortstack{{Speed up selection }$\uparrow$}} \\
        \cmidrule(lr){2-4}
        & PSNR$\uparrow$ & SSIM$\uparrow$ & LPIPS$\downarrow$ & \\
        \midrule
        5
        & 27.518
        & 0.8339
        & 0.1110
        & 1.1$\times$ \\

        10
        & 27.523
        & 0.8467
        & 0.1112
        & 1$\times$ \\

        20
        & 27.538
        & 0.8471
        & 0.1109
        & 0.56$\times$ \\

        50
        & 27.520
        & 0.8463
        & 0.1110
        & 0.34$\times$ \\
        \bottomrule
    \end{tabular}
\caption{Effect of the AGILE-GS+ shortlist candidate view set $N$.}
\label{tab:budget_ablation}
\end{table}



\subsection{Embodied Acquisition in Simulation}
\label{sec:digital-twin}
We evaluate AGILE-GS in a closed-loop embodied setting where the view selector drives a robot arm to reconstruct the map online. The primary platform is an NVIDIA Isaac~Sim~\cite{NVIDIA_Isaac_Sim} digital twin of a robotic lab, built from a textured Scaniverse capture and grounded in real-robot data from a Kinova Gen3 manipulator with an Intel RealSense D435I. Two auxiliary Isaac~Sim scenes are also
evaluated; a compact kitchen with YCB objects on a textured table, and a Warehouse with multiple shelves and reachable YCB grocery items. At each selection event, candidate camera poses are sampled around the workspace and filtered by inverse-kinematics feasibility, joint limits, and
motion continuity. The robot executes the selected trajectory and streams synchronized RGB, depth, and ground-truth pose to an i3DGS~\cite{meuleman2026immediate}
mapper, which incrementally integrates each observation without resetting the model between viewpoints, a property essential for fast, closed-loop map reconstruction.
\begin{figure}[t]
    \centering
    \includegraphics[
        width=0.562\textwidth,
        trim={5.1cm 7.12cm 2.5cm 3.8cm},
        clip
    ]{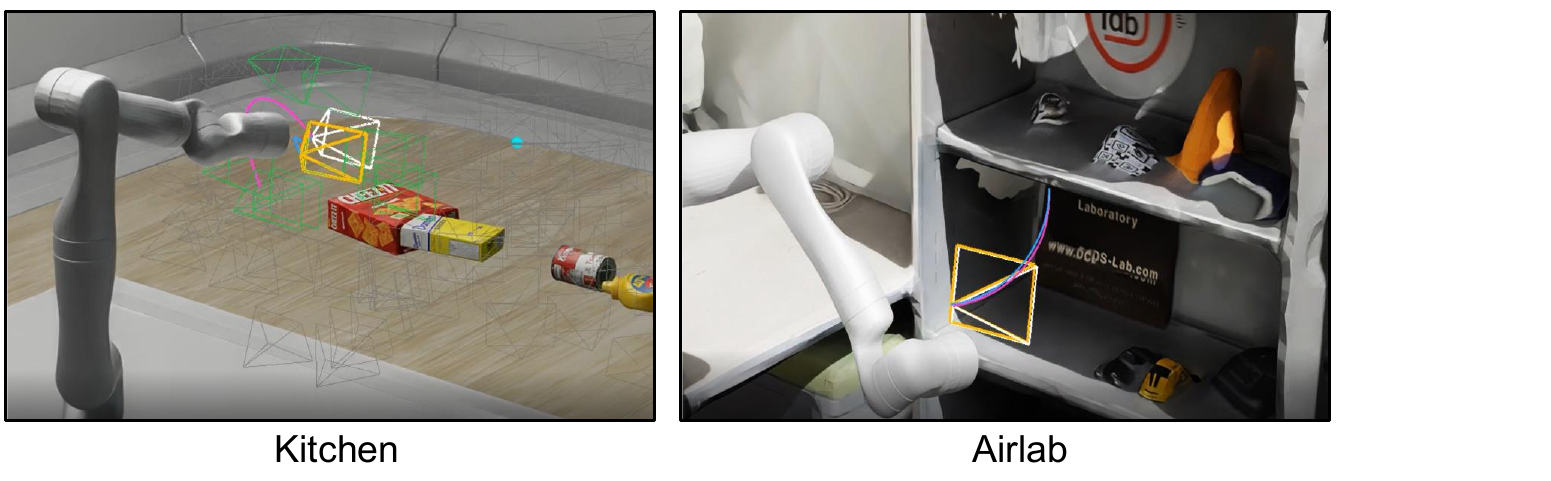}
\vspace{-20pt}
\caption{Embodied selection in Isaac Sim. \textit{Left:} candidate views (green) and the NBV by \textsc{Agile-GS} (yellow).
\textit{Right:} the NBV $T_t^\star$ recovered by Riemannian ascent (yellow), concentrating on the least-constrained region of the scene.}
\vspace{-6pt}
\label{fig:armrobot}
\end{figure}
\begin{table}[t]
\centering
\footnotesize
\setlength{\tabcolsep}{2pt}
\renewcommand{\arraystretch}{0.8}
\begin{tabular}{llcccc}
\toprule
Environment & Method & PSNR$\uparrow$ & SSIM$\uparrow$ & LPIPS$\downarrow$ & Views\,/\,Time (s) \\
\midrule
\multirow{2}{*}{Warehouse}
  & Random   & 17.07 & 0.622 & 0.537 &  \\
  & AGILE-GS  & 18.34 & 0.724 & 0.457 & {9\,/\,144.5 (s)}\\
\midrule
\multirow{2}{*}{Kitchen}
  & Random   & 19.70 & 0.782 & 0.481 &  \\
  & AGILE-GS  & 20.44 & 0.823 & 0.472 & {11\,/\,621.8 (s)}\\
\midrule
\multirow{2}{*}{Airlab}
  & Random   & 21.94 & 0.811 & 0.324 &  \\
  & AGILE-GS  & \textbf{26.36} & \textbf{0.900} & \textbf{0.245} & {100\,/\,998.6 (s)}\\
\bottomrule
\end{tabular}
\vspace{-6pt}
\caption{Embodied reconstruction in Isaac~Sim
environments.}
\vspace{-6pt}
\label{tab:digital-twin}
\end{table}

Table~\ref{tab:digital-twin} compares AGILE-GS against random view
sampling under an identical acquisition budget in each environment.
AGILE-GS improves reconstruction quality over the random baseline. 
This evaluation combines a real-robot simulation Isaac~Sim  with a closed-loop active 3DGS acquisition pipeline, bridging
offline benchmark selection and executable robotic acquisition
under physically consistent online 3DGS mapping and kinematic conditions.



\section{Conclusion}
\label{sec:conclusion}
This work introduced AGILE-GS, a framework for next-best-view
selection in 3D Gaussian Splatting that decouples model-aware
guidance from per-candidate evaluation. A virtual anchor, optimized
on $\mathrm{SE}(3)$ by Riemannian gradient ascent, supplies the
information direction that previously came from Fisher scoring over
the full pool. A fast greedy ridge-leverage step then extracts a
compact, non-redundant proposal set without rendering any candidate.
The shortlist supports two operating modes. AGILE-GS commits to the
leading proposal for minimum latency, and AGILE-GS+ reranks the
shortlist by information gain for higher accuracy. Together they give
a controllable accuracy--efficiency trade-off. Across standard
benchmarks and an embodied evaluation on a manipulator robot, both
modes match or exceed baseline selection at substantially lower cost.
This narrows the gap between offline NBV benchmarks and executable
robotic sensing. Extending the framework to multi-agent coordination
and to safety-constrained online acquisition are natural next steps.

{
    \small
    \bibliographystyle{ieeenat_fullname}
    \bibliography{main}
}

\end{document}